\documentclass[letterpaper, 10 pt, conference]{ieeeconf}  % Comment this line out if you need a4paper

\IEEEoverridecommandlockouts                              % This command is only needed if 
\usepackage{amssymb}  % assumes amsmath package installed
\usepackage{graphicx}
\usepackage{xcolor}
\usepackage{amsmath}
\usepackage{booktabs}
\usepackage{subcaption}
\usepackage{url}

\title{\LARGE \bf
Geometric Entropy: When Trajectory Diversity Helps and Hurts in Imitation Learning
}

\author{Qian Luo$^{1,2}$, Ruizhe Liu$^{1}$, Pei Zhou$^{1,2}$, Xunzhe Zhou$^{1,2}$, and Yanchao Yang$^{1,2}$%
\thanks{$^{1}$Department of Electrical and Electronic Engineering, the University of Hong Kong. $^{2}$Transcengram. e-mail: \{luoqian1,zrllrz360,pezhou,xunzhe\_zhou\}@connect.hku.hk; \mbox{yanchaoy@hku.hk}.}%
}

\begin{document}

\maketitle
\thispagestyle{empty}
\pagestyle{empty}

%%%%%%%%%%%%%%%%%%%%%%%%%%%%%%%%%%%%%%%%%%%%%%%%%%%%%%%%%%%%%%%%%%%%%%%%%%%%%%%%
\begin{abstract}
We study how \emph{trajectory-shape diversity} in demonstrations affects
imitation learning (IL) performance across models, tasks, and data
scales. We introduce \textbf{Geometric Entropy} ($H_G$), a task-agnostic
metric that quantifies intrinsic diversity of transit trajectories after
normalizing away extrinsic variation (e.g., goal pose and workspace
scale) via target-frame alignment.
Across multiple IL architectures and both simulation and real-robot
contact-rich manipulation tasks, we observe a consistent \textbf{inverted-U}
relationship between success and $H_G$: increasing geometric diversity
improves robustness in low-diversity regimes but degrades performance
once diversity induces strategy ambiguity. Moreover, the optimal entropy
shifts toward \emph{lower} values as task mastery increases (via more
data, easier tasks, or stronger priors), and for a pretrained
vision-language-action model the trend becomes effectively monotonic
decreasing. Practically, $H_G$ enables fast pre-training auditing of
demonstration datasets and offers a simple guideline for calibrating
demonstrations toward the learnable regime. Project page:
\url{https://geometric-entropy.github.io/}.
\end{abstract}

%%%%%%%%%%%%%%%%%%%%%%%%%%%%%%%%%%%%%%%%%%%%%%%%%%%%%%%%%%%%%%%%%%%%%%%%%%%%%%%%
\section{INTRODUCTION}
Modern imitation learning (IL) is increasingly data-centric. As policy
classes scale to high-capacity architectures such as diffusion policies
and vision-language-action (VLA) models, the bottleneck often shifts from
algorithm design to \emph{what} demonstrations contain and \emph{how} they
are curated. Prior work studies data quality along axes such as
observation fidelity, label accuracy, and task or scene coverage. Yet a
more fundamental question remains underexplored: \emph{how diverse should
the \textbf{trajectory shapes} in a demonstration dataset be?}

Trajectory-shape diversity is a double-edged sword. On one hand, varied
approach geometries can improve robustness by expanding effective
coverage of relevant state--action regions, reducing overfitting to a
narrow motion corridor. On the other hand, excessive diversity can
introduce \emph{mode inconsistency}: multiple distinct strategies
accomplish the same goal, but the conditioning signal may be insufficient
to reliably select among them. In contact-rich manipulation, even small
confusion (or averaging across incompatible strategies) can yield
infeasible motions and sharp performance drops. This motivates a
non-monotonic relationship between geometric diversity and success: too
little diversity leads to brittleness, while too much becomes
interference.

To study this systematically, we need a pre-training metric that
quantifies \emph{between-trajectory} \textbf{geometric shape diversity}
rather than marginal variability, and that is invariant to extrinsic
episode factors (e.g., target pose, workspace placement, and scale) that
can inflate naive spread measures without reflecting true strategy
differences. We therefore propose \textbf{Geometric Entropy} ($H_G$), a
lightweight, task-agnostic scalar designed to meet these requirements.
$H_G$ isolates the learnable geometric signal by focusing on free-space
\emph{transit} segments and expressing each segment in a canonical,
target-anchored representation: trajectories are arc-length resampled to
a fixed length, translated to the segment endpoint, rotated into the
target frame, and normalized by the start--end displacement to remove
scale. This alignment ensures that trajectories that differ only by where
the task happens map to the same descriptor, while genuine strategy
differences (approach angle, curvature, altitude profile, detours)
remain. $H_G$ then summarizes the intrinsic complexity of the resulting
shape manifold with a stable spectral statistic of the aligned trajectory
descriptors, capturing both the number of independent shape modes and
their overall spread in shape space. Because it is computationally cheap
and sample-efficient, $H_G$ provides a practical audit signal for dataset
curation \emph{before} training.

Using controlled demonstrations with tunable geometric diversity and
evaluating across multiple tasks, models, and data scales in both
simulation and on real robots, we observe a consistent
\textbf{Inverted-U} pattern: performance often improves as $H_G$ increases
from low values, but degrades when $H_G$ becomes too large. Moreover, the
peak shifts toward \textbf{lower} $H_G$ as the learner becomes more
proficient. This peak-shift supports a unifying \textbf{task mastery}
hypothesis: when mastery is low, additional geometric diversity is
beneficial by expanding effective coverage; when mastery is high, the
tolerable diversity shrinks and excessive variation acts as interference
that reduces success.

Our contributions are:
\begin{enumerate}
    \item We provide a \textbf{formal treatment} of trajectory
          \textbf{geometric shape diversity} in imitation learning,
          clarifying why it is a distinct dataset attribute that can be
          beneficial yet harmful beyond a learnable regime.
    \item We propose \textbf{Geometric Entropy} ($H_G$), a practical
          metric that quantifies intrinsic trajectory-shape diversity and
          can be computed prior to training for dataset auditing and
          curation.
    \item Through systematic evaluation across simulation and real-world
          contact-rich manipulation tasks, we report a consistent
          \textbf{Inverted-U} trend and a \textbf{peak-shift} with
          increasing task mastery, providing actionable guidance for
          calibrating demonstration diversity.
\end{enumerate}

The paper is organized as follows. Section~\ref{sec:related} surveys
related work. Section~\ref{sec:method} presents the metric.
Section~\ref{sec:setup} describes the experimental setup, and
Section~\ref{sec:results} reports results and analysis.
Section~\ref{sec:discuss} discusses implications and limitations.

%%%%%%%%%%%%%%%%%%%%%%%%%%%%%%%%%%%%%%%%%%%%%%%%%%%%%%%%%%%%%%%%%%%%%%%%%%%%%%%%
% RELATED WORK
%%%%%%%%%%%%%%%%%%%%%%%%%%%%%%%%%%%%%%%%%%%%%%%%%%%%%%%%%%%%%%%%%%%%%%%%%%%%%%%%
\section{RELATED WORK}
\label{sec:related}

\subsection{Data Scaling in Imitation Learning}
Recent robotics imitation learning has shifted towards scaling, with policies and agents improving via large, heterogeneous robot datasets and multi-embodiment trajectories, as seen in RT-1~\cite{brohan2022rt}, RT-2~\cite{zitkovich2023rt}, Open X-Embodiment/RT-X~\cite{o2024open}, DROID~\cite{khazatsky2024droid}, Octo~\cite{octo_2023}, and OpenVLA~\cite{kim24openvla}. 
Efforts like BC-Z~\cite{jang2022bc} and RoboCat~\cite{bousmalis2023robocat} further advocate for broad generalization through scaling task diversity, while embodied multimodal models bridge internet-scale pretraining with robot control~\cite{driess2023palm}. 
In parallel, simulation-scale imitation benchmarks and prompted generalist manipulation agents operationalize scaling along tasks and prompts~\cite{jiang2023vima}.
Hybrid offline pipelines~\cite{lu2022aw}, show the importance of large-scale experience combined with strong policy architectures. 
However, these approaches typically treat ``diversity'' as a coarse dataset attribute (tasks/objects/embodiments/modalities) without focusing on trajectory geometry, a gap our work addresses through geometric entropy and an IL-specific inverted-U trend that indicates when geometric diversity aids or hinders learning.

\subsection{Data Curation and Collection for Imitation Learning}
Beyond scale, recent work formalizes IL data quality and develops curation/collection mechanisms. Data quality in IL decomposes performance under distribution shift into action divergence and transition diversity~\cite{belkhale2023data}; interactive IL such as DAgger collects labels on the learner-induced distribution to mitigate compounding error~\cite{ross2011reduction}, with related robustness via noise-injected supervision~\cite{laskey2017dart}. Offline learning is also sensitive to demonstration quality~\cite{mandlekar2022matters}, motivating reweighting/filtering from suboptimal data~\cite{xu2022discriminator}, transition-level self-supervised curation~\cite{zhang2026scizor}, influence-based attribution to closed-loop performance~\cite{agia2025cupid}, and online experience-guided demo selection~\cite{chen2025curating}, as well as learning with imperfect demonstrations via self-supervision~\cite{wu2025learning}. Complementary directions synthesize data or inject variability to expand coverage/occupancy, including MimicGen~\cite{mandlekar2023mimicgen}, DemoGen~\cite{xuedemogen}, ManiBox~\cite{tan2024manibox}, adversarial perturbation~\cite{huang2025adversarial}, FieldGen~\cite{wang2025fieldgen}, and MOVE~\cite{wang2025move}. Unlike methods that primarily optimize selection, robustness, or coverage, we directly diagnose trajectory geometry as an intrinsic dataset property and provide a predictive metric plus an inverted-U trend to guide whether to increase or reduce geometric diversity.

\subsection{Trajectory Geometry, Diversity, and Mode Coverage}
Trajectory diversity is commonly studied as multimodality/mode coverage in conditional action distributions: multi-modal IL learns stochastic multi-skill policies~\cite{hausman2017multi}, and modern sequence/generative policies model multimodality via discretization or generation~\cite{shafiullah2022behavior,chi2025diffusion}, with related formulations including implicit energy-based action modeling~\cite{florence2022implicit}, hierarchical tokenization~\cite{lee2024behavior}, and efficient multi-task transformers~\cite{haldar2024baku}. Diversity is also evaluated via benchmarks (D3IL)~\cite{jiatowards} and distributional views such as occupancy-measure matching~\cite{ho2016generative}
, while analyses show its effect depends on which factors vary and can even hinder learning~\cite{shi2025diversity}
; orthogonally, classical sequence-alignment tools quantify pairwise trajectory similarity/structure~\cite{sakoe2003dynamic}.
% Procrustes~\cite{gower1975generalized}, elastic shape metrics~\cite{srivastava2010shape}, 
% persistent homology~\cite{pokorny2016topological,bhattacharya2015persistent}). 
However, these lines do not provide an IL-specific, dataset-level criterion for \emph{how much} geometric diversity is learnable under different priors; we address this with extrinsic-invariant geometric entropy and a predictive inverted-U trend (with peak shifts across priors) for curation.

%%%%%%%%%%%%%%%%%%%%%%%%%%%%%%%%%%%%%%%%%%%%%%%%%%%%%%%%%%%%%%%%%%%%%%%%%%%%%%%%

% METHODOLOGY
%%%%%%%%%%%%%%%%%%%%%%%%%%%%%%%%%%%%%%%%%%%%%%%%%%%%%%%%%%%%%%%%%%%%%%%%%%%%%%%%
\section{METHODOLOGY}
\label{sec:method}

We seek a metric that answers a single question: given a set of
demonstrations for a manipulation task, how geometrically diverse are the
trajectory \emph{shapes}, independent of where in the workspace the task
happens to take place? This requires three components: a formal
decomposition of the factors that shape raw trajectories
(Section~\ref{sec:formulation}), a phase-aware alignment pipeline that
isolates the geometric signal (Sections~\ref{sec:phase}
and~\ref{sec:alignment}), and a spectral diversity measure robust to the
high-dimensional, moderate-sample regime of robotics
(Section~\ref{sec:entropy}).

% ─────────────────────────────────────────────────────────────────────
\subsection{Problem Formulation}
\label{sec:formulation}

Let a demonstration dataset $\mathcal{D} = \{\tau_1, \dots, \tau_M\}$
consist of $M$ trajectories, where each trajectory
$\tau_i = \{\mathbf{p}_1^{(i)}, \dots, \mathbf{p}_{L_i}^{(i)}\}$ is a
variable-length sequence of end-effector positions
$\mathbf{p}_t \in \mathbb{R}^3$. In a typical manipulation episode, the
raw positions are shaped by three entangled factors:

\begin{itemize}
\item \textbf{Extrinsic context $\mathcal{E}$}: the object/goal \emph{pose}
(position and orientation) and workspace scale, which vary across
episodes due to domain randomization or different environment
configurations. These factors determine \emph{where} and \emph{how} the
task is situated in the workspace but carry no information about the
demonstrator's motion strategy. Our alignment normalizes this context by
expressing trajectories in a target-anchored local frame and removing
scale (Section~\ref{sec:alignment}).

\item \textbf{Geometric strategy $\mathcal{G}$}: the intrinsic shape of
      the motion path---its approach angle, curvature profile, altitude
      trajectory, and detour geometry---which reflects the demonstrator's
      choice of \emph{how} to reach the target. This is the quantity
      whose diversity we wish to measure.
\item \textbf{Execution noise $\epsilon$}: controller jitter, motion
      planner non-determinism, and other sources of irreproducibility
      that add stochastic variation unrelated to either context or
      strategy.
\end{itemize}

The central challenge is to quantify the diversity of $\mathcal{G}$
across the dataset while marginalizing away $\mathcal{E}$ and $\epsilon$.
Failure to separate these factors leads to inflated diversity estimates:
a dataset of identical straight-line approaches executed at randomized
goal locations will appear highly diverse under any metric that operates
on raw positions, yet it contains zero genuine strategy variation. Our
pipeline addresses this by first decomposing episodes into semantically
meaningful segments (Section~\ref{sec:phase}), then applying a
normalization that strips extrinsic context while preserving the geometric
signal (Section~\ref{sec:alignment}).

% ─────────────────────────────────────────────────────────────────────
\subsection{Phase Decomposition: Transit vs.\ Fine Manipulation}
\label{sec:phase}

A manipulation episode typically interleaves two qualitatively different
motion regimes:

\begin{itemize}
\item \textbf{Transit motion}: free-space movement of the end-effector
      toward (or away from) a task-relevant target. Its geometry---the
      approach angle, curvature, and altitude profile---is the primary
      carrier of demonstrator strategy diversity and the locus of
      learnable geometric variation.
\item \textbf{Fine manipulation}~\cite{wang2025fieldgen}: contact-rich actions such as grasping,
      insertion, pivoting, and release, where the end-effector trajectory
      is tightly constrained by object geometry and contact mechanics,
      leaving little room for meaningful geometric variation.
\end{itemize}

We focus exclusively on transit segments. Phase boundaries are identified
by task-specific semantic events: for example, gripper closure marks the
transition from approach to grasp, and gripper opening marks the
transition from transport to release. This decomposition generalizes to
any multi-phase task---a pick-and-place episode yields pick-transit and
place-transit segments, while a more complex assembly task might yield
three or more transit phases separated by distinct contact events.

Each transit segment is aligned and scored independently
(Sections~\ref{sec:alignment}--\ref{sec:entropy}). This per-phase
treatment is essential: mixing transit and fine-manipulation segments
would conflate genuine strategy diversity with contact-constrained motion,
diluting the diagnostic power of the metric.

% ─────────────────────────────────────────────────────────────────────
\subsection{Endpoint-Anchored Frame Alignment}
\label{sec:alignment}

We transform each transit trajectory into a canonical shape descriptor
$\boldsymbol{\xi} \in \mathbb{R}^{3T}$ via four steps designed to remove
extrinsic variation while preserving intrinsic geometric strategy.

\textbf{Step~1: Arc-length resampling.}
Raw trajectories vary in length due to differences in execution speed and
controller frequency. We resample each trajectory to exactly $T$
waypoints (We set $T{=}50$ in our experiments) at uniform arc-length intervals via linear interpolation,
ensuring that the descriptor captures \emph{where} the end-effector goes
but not \emph{when} it arrives, and that all trajectories live in the
same fixed-dimensional space $\mathbb{R}^{3T}$.

\textbf{Step~2: Endpoint anchoring (translation normalization).}
We translate each resampled trajectory so that its terminal point---the
task target (e.g., the grasp pose)---sits at the origin:
$\mathbf{p}'_k = \bar{\mathbf{p}}_k - \bar{\mathbf{p}}_T$.
Anchoring at the terminal point cancels extrinsic positional offsets due
to randomized target locations.

\textbf{Step~3: Target-frame alignment (rotation normalization).}
To remove extrinsic rotational variation, we express the anchored
trajectory in a canonical local frame attached to the task target. Let
$\mathbf{R}_T \in SO(3)$ denote the target (or object) orientation used
to define the local task frame for the corresponding phase.\footnote{In
simulation, $\mathbf{R}_T$ is available from environment state; on real
robots, it is obtained from the same perception/state estimation used to
define the task target.} We rotate the anchored points into this frame:
$\mathbf{p}''_k = \mathbf{R}_T^\top \mathbf{p}'_k$.
This step ensures that trajectories that are identical up to a global
rotation (e.g., due to different object yaw) map to the same shape
descriptor.

\textbf{Step~4: Displacement scaling.}
We normalize by the Euclidean distance between the start and end of the
segment:
$\tilde{\mathbf{p}}_k = \mathbf{p}''_k \,/\,
(\|\bar{\mathbf{p}}_1 - \bar{\mathbf{p}}_T\| + \epsilon)$,
with $\epsilon = 10^{-9}$. This removes workspace-scale effects so that a
short reach and a long reach with the same curvature profile yield
identical descriptors---we measure \emph{shape}, not physical extent.

The canonical descriptor is the flattened vector
$\boldsymbol{\xi} = [\tilde{\mathbf{p}}_1^\top, \dots,
\tilde{\mathbf{p}}_T^\top]^\top \in \mathbb{R}^{3T}$.
After this pipeline, two trajectories produce the same $\boldsymbol{\xi}$
if and only if they share the same intrinsic shape relative to the target
frame (up to residual execution noise).

% ─────────────────────────────────────────────────────────────────────
\subsection{Geometric Entropy ($H_G$)}
\label{sec:entropy}

Given $M$ aligned descriptors
$\{\boldsymbol{\xi}_1,\dots,\boldsymbol{\xi}_M\}\subset\mathbb{R}^{3T}$,
we seek a single scalar that summarizes geometric shape diversity. In our
setting, the descriptor dimension is moderately high ($3T{=}150$) while
demonstration counts are typically $M\sim 10^2$--$10^3$, where fully
non-parametric density estimation can be unreliable. We therefore
summarize diversity through the covariance eigenspectrum, and later
empirically compare this choice against common alternatives in
Section~\ref{sec:metric_comparison}.

\textbf{Spectral decomposition.}
To avoid estimating a probability density in a high-dimensional descriptor
space ($3T{=}150$) with moderate sample sizes ($M\sim 10^2$--$10^3$), we
summarize diversity through the covariance eigenspectrum. Let
$\mathbf{X}\in\mathbb{R}^{M\times 3T}$ be the mean-centered descriptor
matrix. Its SVD yields eigenvalues $\{\lambda_j\}$ of the sample
covariance, where each $\lambda_j$ captures variance along one
orthogonal shape mode.

A useful diversity metric should capture two factors: \emph{how many}
independent shape modes exist (structural complexity) and \emph{how
spread out} they are (variation magnitude). We quantify the former via
the \textbf{effective dimensionality} $d_{\mathrm{eff}}$, the
exponentiated Shannon entropy of the normalized spectrum
$p_j=\lambda_j/\sum_k \lambda_k$:
\begin{equation}
d_{\mathrm{eff}}=\exp\!\Bigl(-\sum_{j:\,p_j>0} p_j\ln p_j\Bigr).
\label{eq:deff}
\end{equation}
When variance concentrates on a single mode, $d_{\mathrm{eff}}{=}1$; when
it is uniform across $D$ modes, $d_{\mathrm{eff}}{=}D$.

We then define the \textbf{Geometric Entropy}:
\begin{equation}
H_G=\ln d_{\mathrm{eff}}+\ln\!\Bigl(\sum_j \lambda_j\Bigr).
\label{eq:hg}
\end{equation}
The first term increases as the dataset populates more independent
approach strategies; the second increases as these modes spread further
apart in aligned shape space. Both are necessary: many modes separated by
tiny angles yield high $d_{\mathrm{eff}}$ but negligible magnitude,
whereas a single dominant mode spanning a large arc yields low
$d_{\mathrm{eff}}$ but high magnitude. Unlike $\log\det(\Sigma)$, which
is ill-conditioned under $M \ll 3T$ and requires ad-hoc regularization,
the total-variance term $\log\!\big(\sum_j\lambda_j\big)$ remains stable
in our regime.

Rather than claiming a strict approximation of differential entropy, we
use $H_G$ as a \textbf{lightweight spectral surrogate}: it is
non-parametric (no bandwidth to tune), computationally cheap (a single
SVD), and well-behaved at practical $M$. We later contrast $H_G$ with
representative baselines in Section~\ref{sec:metric_comparison}.

\textbf{Multi-phase aggregation.}
For tasks with $P$ transit phases, we compute $H_G$ per phase and define
the overall dataset entropy as the size-weighted average:
\begin{equation}
H_G^{\mathrm{avg}}=
\frac{\sum_{p=1}^{P} n_p \cdot H_G^{(p)}}{\sum_{p=1}^{P} n_p},
\label{eq:hg_avg}
\end{equation}
where $n_p$ is the number of successful trajectories for phase $p$.

%%%%%%%%%%%%%%%%%%%%%%%%%%%%%%%%%%%%%%%%%%%%%%%%%%%%%%%%%%%%%%%%%%%%%%%%%%%%%%%%
% EXPERIMENTAL SETUP
%%%%%%%%%%%%%%%%%%%%%%%%%%%%%%%%%%%%%%%%%%%%%%%%%%%%%%%%%%%%%%%%%%%%%%%%%%%%%%%%
%%%%%%%%%%%%%%%%%%%%%%%%%%%%%%%%%%%%%%%%%%%%%%%%%%%%%%%%%%%%%%%%%%%%%%%%%%%%%%%%
% EXPERIMENTAL SETUP
%%%%%%%%%%%%%%%%%%%%%%%%%%%%%%%%%%%%%%%%%%%%%%%%%%%%%%%%%%%%%%%%%%%%%%%%%%%%%%%%
\section{EXPERIMENTAL SETUP}
\label{sec:setup}

Our experiments are designed to answer three questions:
(1)~Is $H_G$ a stable and discriminative descriptor of trajectory-shape
diversity?
(2)~How does imitation learning performance evolve as a function of $H_G$
across tasks, models, and data scales?
(3)~Can we identify an organizing principle that governs the optimal
diversity level?

To isolate the effect of \emph{geometric strategy diversity} from other
factors, we require datasets whose trajectory shapes can be varied
\emph{continuously and independently} while holding task definition and
observation/action interfaces fixed. We therefore build a parameterized
demonstration generator with controllable diversity and then validate the
observed trends on real-robot hardware.

% ─────────────────────────────────────────────────────────────────────
\subsection{Simulation Experiments}
\label{sec:sim}

\subsubsection{Parameterized Demonstration Generation}

\begin{figure}[thpb]
      \centering
      \includegraphics[width=0.48\textwidth]{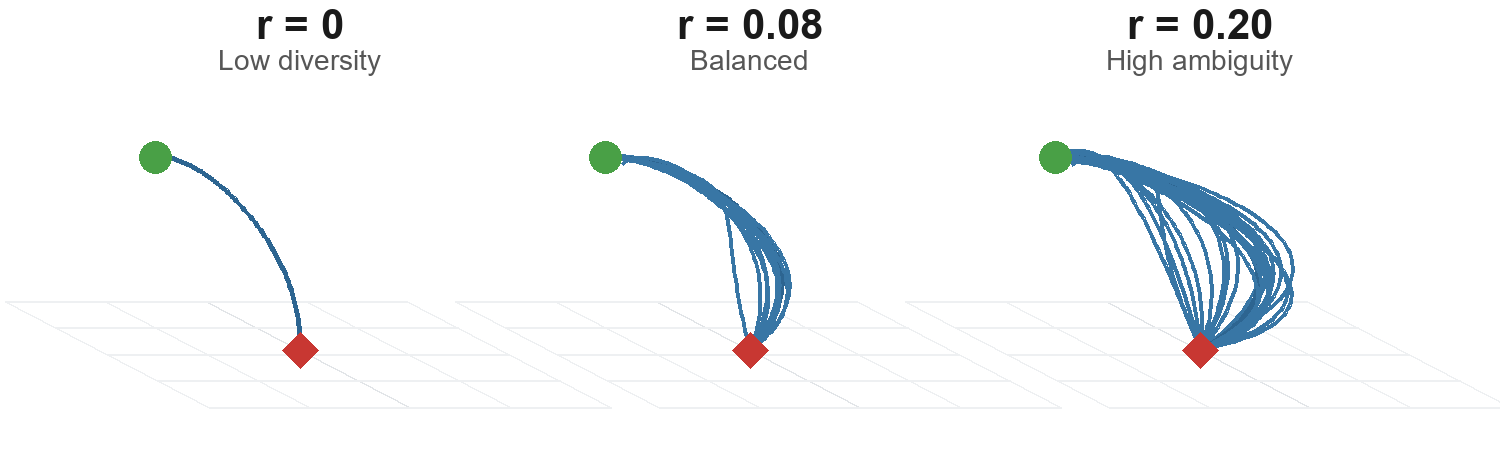}
      \caption{Trajectory manifolds under varying approach radius $r$.
      At $r \approx 0$ (left), demonstrations follow near-identical
      descents. At moderate $r$ (center), approach paths form a cone of
      curved arcs. At large $r$ (right), the manifold expands further and
      can enter a regime where excessive geometric diversity induces mode
      ambiguity that degrades learning.}
      \label{fig:schematic}
\end{figure}

Our simulation pipeline is built on the \textbf{PandaArmMotionPlanningSolver}
from ManiSkill3~\cite{tao2024maniskill3}. We inject controlled geometric
diversity by introducing an \emph{approach waypoint} that the end-effector
must pass through before reaching the task target. Two parameters control
diversity:

\textbf{Approach radius $r$.}
For each demonstration, we sample a waypoint on a circle of radius $r$
centered above the target:
\begin{equation}
\mathbf{w} = \mathbf{g} +
  [r\cos\theta,\; r\sin\theta,\; \Delta z]^\top,
\label{eq:waypoint}
\end{equation}
where $\mathbf{g}$ is the target position, $\theta$ is the approach
angle, and $\Delta z$ is a fixed height offset. When $r{=}0$,
trajectories are near-identical straight descents; larger $r$ yields
increasingly curved, laterally displaced approaches
(Fig.~\ref{fig:schematic}).

\textbf{Discrete mode count $k$.}
We sample $\theta \sim \mathrm{Uniform}(0,2\pi)$ for $k{=}0$, or from $k$
equally spaced discrete values for $k{>}0$. This decouples
\emph{magnitude} from \emph{modality}: large $r$ with $k{=}1$ produces
long but consistent (single-strategy) approaches, while small $r$ with
$k{=}0$ produces short but angularly scattered trajectories. This
controlled separation lets us distinguish magnitude-driven variation from
modality-driven ambiguity in Section~\ref{sec:results}.

For each $(r, k)$ configuration, we generate datasets of size
$N \in \{100, 200, 500, 1000\}$, segment each episode into transit phases
based on gripper events, and compute $H_G$ per phase, averaged using
Eq.~\ref{eq:hg_avg}. All simulation results are reported over three
independent seeds.

\subsubsection{Convergence and Stability of $H_G$}
\label{sec:stability}

\begin{figure}[t]
  \centering
  \includegraphics[width=0.48\textwidth]{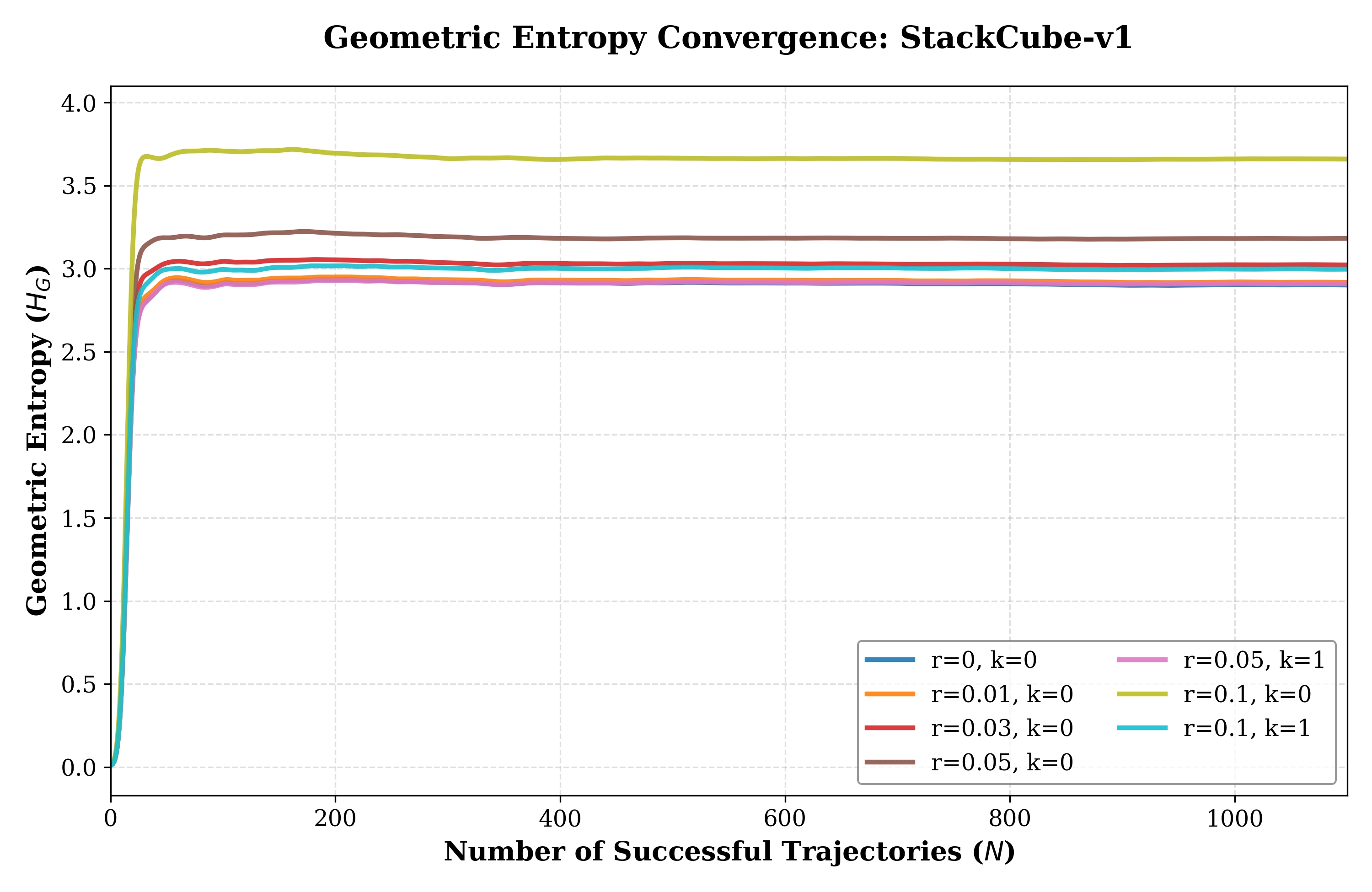}
  \caption{$H_G$ convergence on \texttt{StackCube-v1}. Across all $(r, k)$
  settings, $H_G$ stabilizes within the first $50$--$100$ successful
  trajectories, enabling reliable pre-training audits at practical
  collection scales.}
  \label{fig:convergence}
  \vspace{-0.7em}
\end{figure}

A practical dataset-audit metric should be sample-efficient and stable.
Fig.~\ref{fig:convergence} plots $H_G$ on \texttt{StackCube-v1} as a
function of the number of successful trajectories $N$ across all $(r, k)$
configurations. In all cases, $H_G$ stabilizes within the first
$50$--$100$ demonstrations; for $N \gtrsim 100$, residual fluctuations are
small compared to gaps between configurations.

To make ordering explicit, Table~\ref{tab:hg_values} reports converged
$H_G^{\mathrm{avg}}$ (Eq.~\ref{eq:hg_avg}). For $k{=}0$, $H_G$ increases
monotonically with $r$, confirming that enlarging the waypoint radius
expands the aligned shape manifold. Enforcing a single approach mode
($k{=}1$) substantially lowers $H_G$ even at large $r$, reflecting reduced
mode ambiguity. The non-zero baseline at $(r,k){=}(0,0)$ captures residual
variation from environment randomization and execution noise.

\begin{table}[t]
\centering
\scriptsize
\setlength{\tabcolsep}{3pt}
\renewcommand{\arraystretch}{1.05}
\caption{Converged $H_G^{\mathrm{avg}}$ on \texttt{StackCube-v1}.}
\vspace{-0.5em}
\label{tab:hg_values}
\resizebox{\columnwidth}{!}{%
\begin{tabular}{c|ccccccc}
\toprule
$r/k$ & $0/0$ & $0.01/0$ & $0.03/0$ & $0.05/0$ & $0.05/1$ & $0.10/0$ & $0.10/1$ \\
\midrule
$H_G^{\mathrm{avg}}$ & 2.901 & 2.917 & 3.016 & 3.183 & 2.909 & 3.667 & 2.997 \\
\bottomrule
\end{tabular}%
}
\vspace{-0.5em}
\end{table}

\subsubsection{Task Environments}

We evaluate two ManiSkill3 tasks spanning distinct manipulation regimes:

\textbf{StackCube-v1}: pick up a cube and stack it on a second cube.
Success requires robust grasping under geometric variation and precise
placement alignment.

\textbf{PegInsertionSide-v1}: laterally insert a peg into a hole with
sub-centimeter clearance. This high-precision task reaches near-saturation
performance at moderate $N$, providing a ``high mastery'' contrast to
StackCube.

\subsubsection{Imitation Learning Models}

We evaluate two architectures spanning the prior-knowledge spectrum:

\textbf{Diffusion Policy (DP)~\cite{chi2025diffusion}}: a specialist
model trained from scratch on each dataset.

\textbf{$\pi_{0.5}$~\cite{intelligence2025pi_}}: a vision-language-action
(VLA) model developed by \textbf{Physical Intelligence}, fine-tuned on
StackCube-v1 in our VLA study. Its large-scale pre-training provides a strong prior that we
hypothesize shifts the optimal diversity toward lower $H_G$.

% ─────────────────────────────────────────────────────────────────────
\subsection{Real-Robot Experiments}
\label{sec:real}

To verify that simulation trends are not artifacts of simulated dynamics,
we run real-robot experiments on an ARX robot arm~\cite{arx} across three tasks:
\textbf{StackCube}, \textbf{PlacePanda} (place a toy panda into a basket),
and \textbf{OpenDrawer}. Demonstrations are collected at five diversity
levels by three collectors, with $N{=}100$ demonstrations per
configuration. We compute $H_G$ using the same alignment pipeline and
train \textbf{ACT}~\cite{zhao2023learning} policies. Success is evaluated
over 40 test episodes per configuration. All real-robot results are reported
over three independent seeds.

%%%%%%%%%%%%%%%%%%%%%%%%%%%%%%%%%%%%%%%%%%%%%%%%%%%%%%%%%%%%%%%%%%%%%%%%%%%%%%%%
%%%%%%%%%%%%%%%%%%%%%%%%%%%%%%%%%%%%%%%%%%%%%%%%%%%%%%%%%%%%%%%%%%%%%%%%%%%%%%%%
% RESULTS AND ANALYSIS
%%%%%%%%%%%%%%%%%%%%%%%%%%%%%%%%%%%%%%%%%%%%%%%%%%%%%%%%%%%%%%%%%%%%%%%%%%%%%%%%
\section{RESULTS AND ANALYSIS}
\label{sec:results}

We analyze how \emph{trajectory geometric shape diversity} impacts
imitation learning performance. Across models, tasks, and data scales,
the effect of diversity is not fixed: it depends on the learner's
proficiency, which we summarize as \emph{task mastery}.

% ─────────────────────────────────────────────────────────────────────
\subsection{Simulation Results}
\label{sec:sim_results}

\begin{figure}[t]
  \centering
  \begin{subfigure}{\linewidth}
    \centering
    \includegraphics[width=0.85\linewidth]{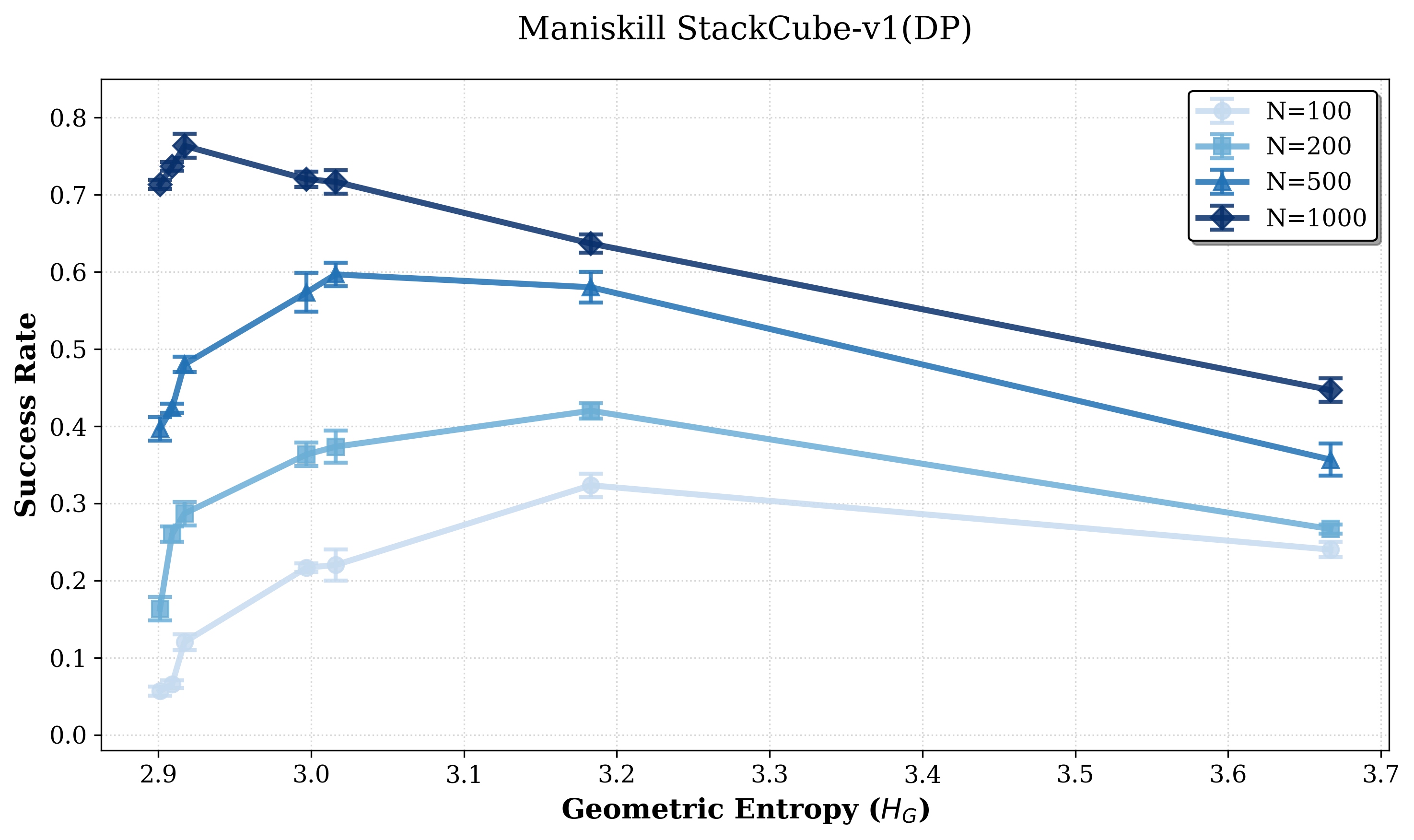}
    \caption{StackCube-v1}
    \label{fig:res_stack}
  \end{subfigure}
  \\[1ex]
  \begin{subfigure}{\linewidth}
    \centering
    \includegraphics[width=0.85\linewidth]{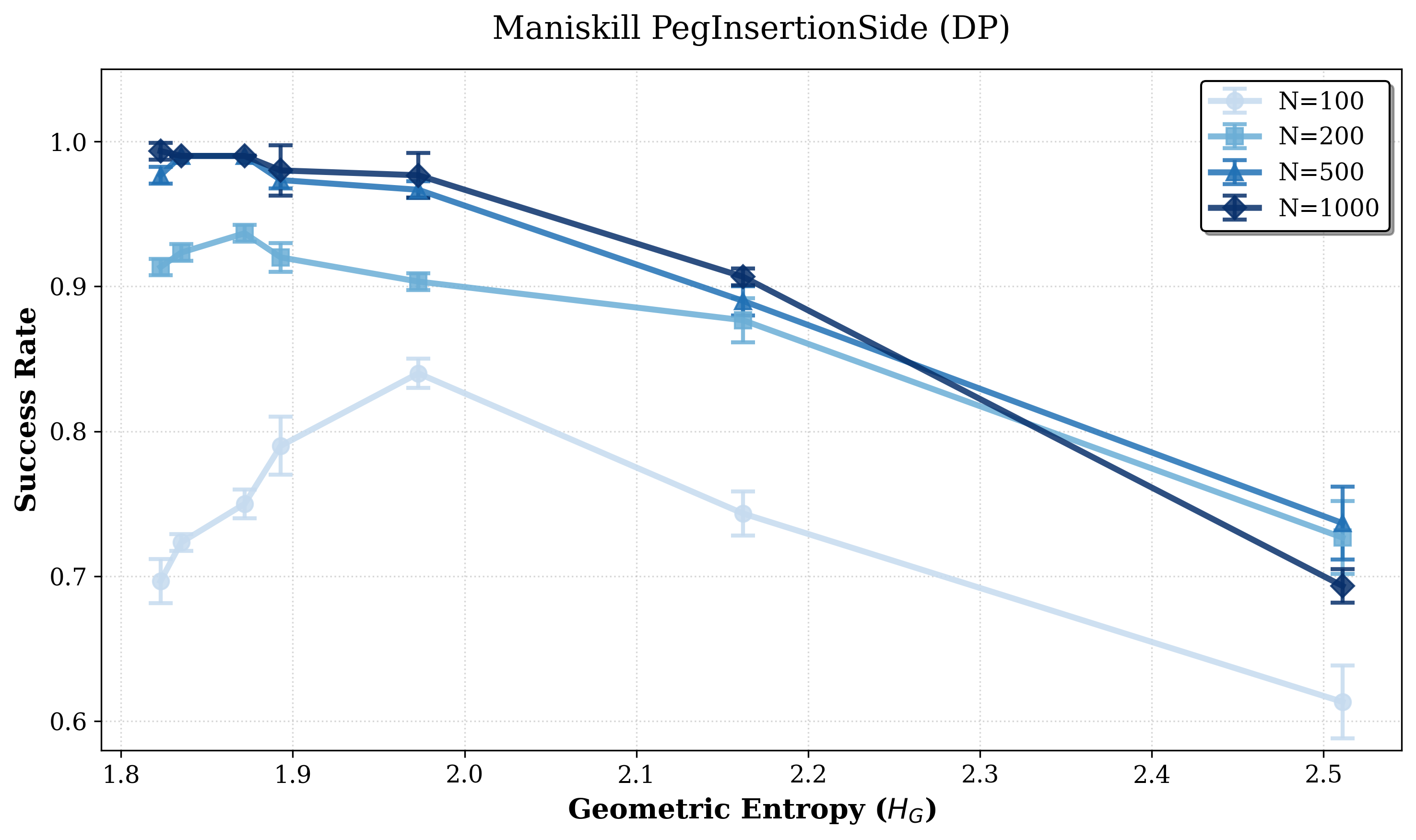}
    \caption{PegInsertionSide-v1}
    \label{fig:res_peg}
  \end{subfigure}
  \caption{Diffusion Policy success rate vs.\ Geometric Entropy ($H_G$).
  Both tasks exhibit an \textbf{Inverted-U} trend and a \textbf{peak-shift}
  toward lower $H_G$ as dataset size $N$ increases. The easier
  \texttt{PegInsertion} task saturates faster than \texttt{StackCube},
  consistent with the mastery hypothesis. Error bars denote variation
  over three seeds.}
  \label{fig:main_dp_results}
  \vspace{-0.7em}
\end{figure}

\textbf{Diffusion Policy: Inverted-U Pattern and Peak-Shift.}
Fig.~\ref{fig:main_dp_results} reports Diffusion Policy performance on
both tasks as a function of $H_G$. Two consistent effects emerge:

\begin{enumerate}
\item \textit{Inverted-U empirical pattern.} Performance often improves
as $H_G$ increases from low values, but degrades once $H_G$ becomes too
large. At low entropy, demonstrations occupy a narrow motion corridor;
when test-time initial conditions deviate, the policy lacks coverage to
recover. Increasing $H_G$ expands geometric coverage and improves
robustness. Beyond a task- and model-dependent peak $H_G^*$, however,
demonstrations contain multiple qualitatively different strategies for
the same conditioning signal, introducing \emph{mode ambiguity}: the
learner may execute an incorrect mode or interpolate between incompatible
strategies, producing infeasible motions—especially near contact.

\item \textit{Peak-shift with data volume.} The optimal $H_G^*$ shifts
leftward as dataset size $N$ increases. With more demonstrations, the
training distribution covers a wider range of initial conditions even
without injected geometric diversity, reducing the marginal benefit of
higher $H_G$ while the cost of ambiguity persists. Larger datasets thus
favor more consistent demonstrations.
\end{enumerate}

\textbf{Comparison across tasks.}
The contrast between tasks highlights how difficulty interacts with
diversity. \texttt{StackCube-v1} remains challenging even at large $N$;
here, moderate increases in $H_G$ are beneficial because added coverage
directly improves robustness. \texttt{PegInsertionSide-v1} reaches
near-saturation at moderate $N$; once the motion corridor is learned,
additional geometric diversity provides little benefit and primarily
acts as interference, leading to a sharper post-peak drop. This supports
the view that higher mastery reduces the tolerable diversity range.

\begin{figure}[thpb]
      \centering
      \includegraphics[width=0.45\textwidth]{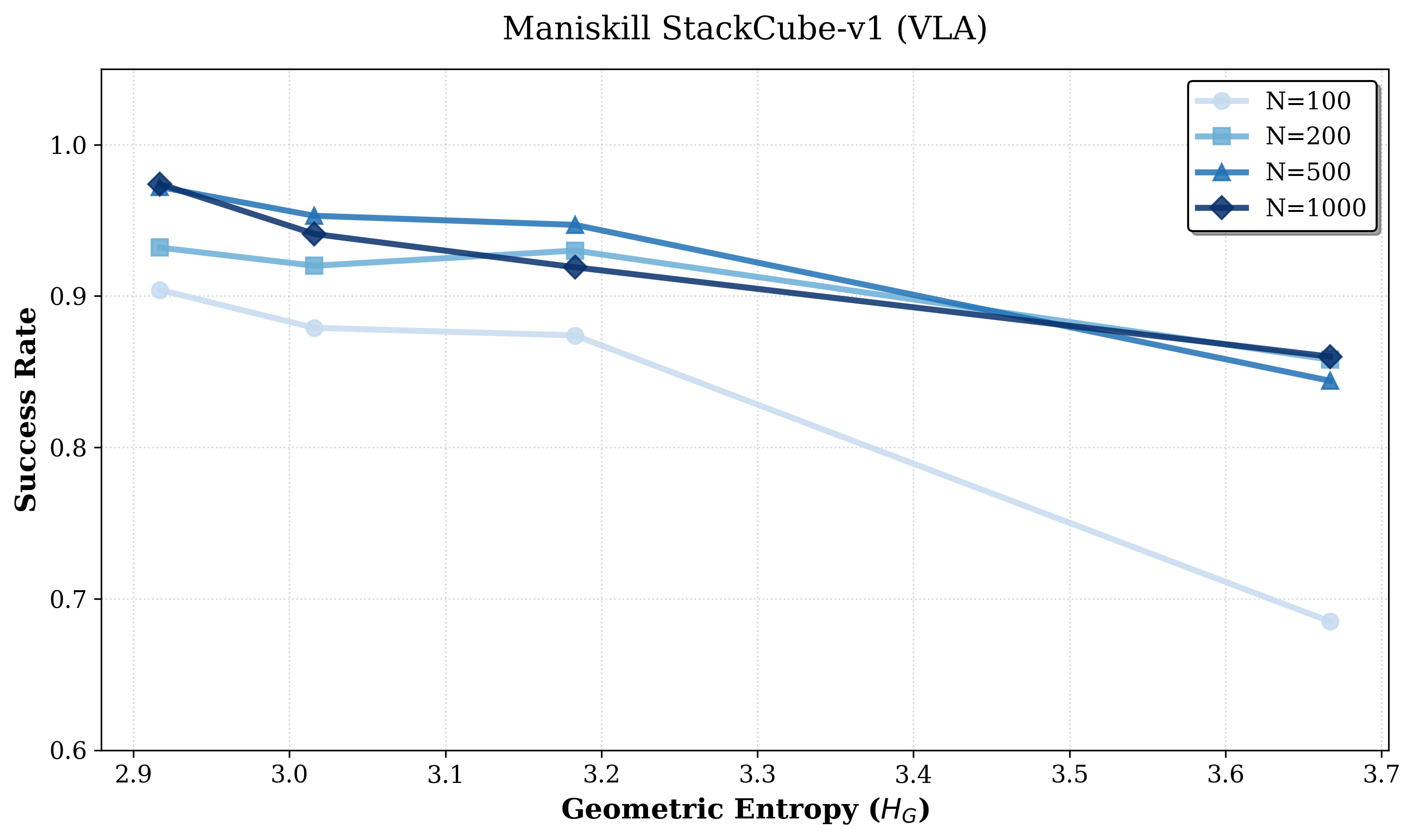}
      \caption{$\pi_{0.5}$ success rate vs.\ $H_G$ on StackCube-v1. The
      approximately monotonically decreasing curve is consistent with a
      high-mastery regime where additional geometric diversity acts as
      interference.}
      \label{fig:vla_res}
      \vspace{-0.7em}
\end{figure}

\textbf{Generalist foundation model ($\pi_{0.5}$): monotonic decay.}
We fine-tune $\pi_{0.5}$ on \texttt{StackCube-v1} (Fig.~\ref{fig:vla_res}). When
plotted against $H_G$, performance decreases approximately monotonically.
This is consistent with a high-mastery regime induced by strong
pretraining: the model already brings substantial geometric coverage, so
additional diversity in the fine-tuning set is redundant and can be
interpreted as conflicting behavior, degrading execution of the motion
corridors preferred by the prior. In effect, the Inverted-U peak has
shifted so far left that no beneficial diversity regime is visible within
the tested range.

% ─────────────────────────────────────────────────────────────────────
\subsection{The Mastery--Entropy Principle}
\label{sec:mastery}

Synthesizing results across tasks, data scales, and model architectures
suggests an organizing principle: the optimal geometric entropy $H_G^*$ is
governed by the learner's \textbf{task mastery}. We use \emph{task mastery}
to denote how proficient the learner is relative to a task's demands,
shaped by (i) training data volume/coverage ($N$), (ii) task difficulty,
and (iii) model priors (pretraining and inductive biases). As mastery
increases, the learner requires less geometric diversity to achieve
robustness, and tolerance for conflicting modes shrinks.

\paragraph{Operationalizing mastery.}
Task mastery is a latent notion rather than a single measured variable,
but it can be approximated with simple proxies. A direct proxy is the
success rate achieved when training on a \emph{low-diversity} (or
``no-injected-diversity'') demonstration set under the same evaluation
protocol. High baseline success indicates that the learner already
captures the relevant motion corridor (high mastery), so additional
geometric diversity is more likely to act as interference; low baseline
success suggests insufficient coverage (low mastery), where increasing
diversity can improve robustness. Increasing $N$, reducing task
difficulty, or strengthening priors typically increases this baseline and
shifts the optimal $H_G^*$ toward lower values.

\begin{itemize}
\item \textbf{Low mastery} (e.g., \texttt{StackCube} at small $N$, or a
      model trained from scratch): moderate increases in $H_G$ help by
      expanding effective coverage; the peak lies at higher $H_G$.
\item \textbf{High mastery} (e.g., \texttt{PegInsertion} at large $N$, or
      a pre-trained VLA): consistent demonstrations already suffice;
      additional diversity is primarily interference, and the relationship
      becomes effectively monotonic decreasing.
\end{itemize}

% ─────────────────────────────────────────────────────────────────────
\subsection{Real-Robot Validation}
\label{sec:real_results}

\begin{figure}[tb]
  \centering
  \includegraphics[width=\columnwidth]{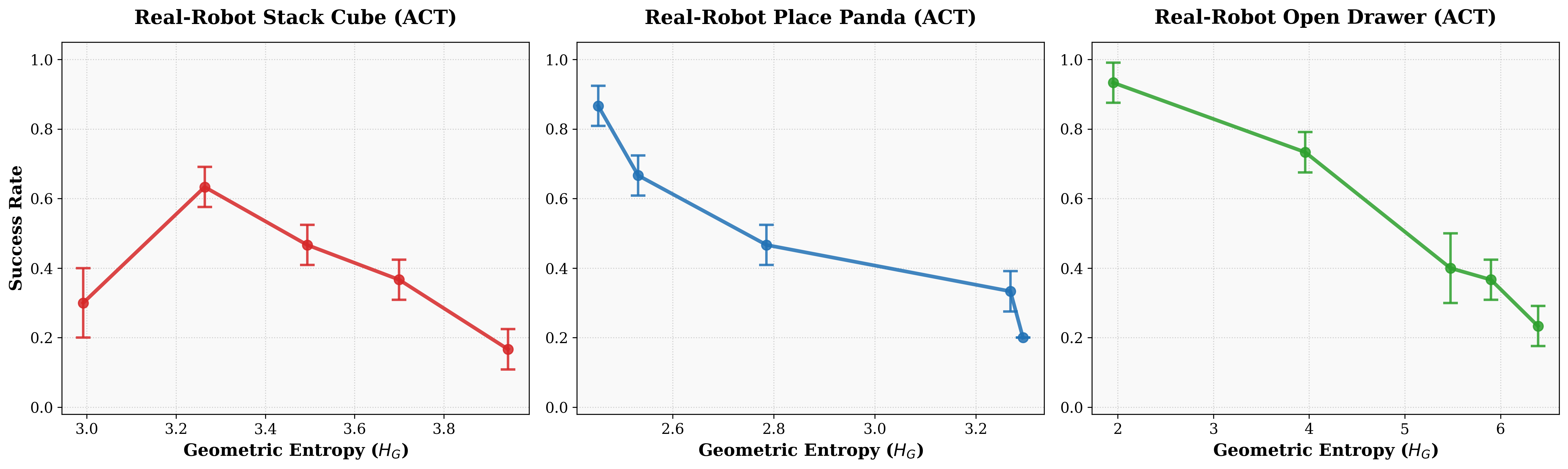}
  \caption{Real-robot success rate vs.\ $H_G$ for three tasks. StackCube
  (left) exhibits a clear Inverted-U, peaking at $H_G \approx 3.3$.
  PlacePanda (center) and OpenDrawer (right) show monotonic decline,
  consistent with high-mastery tasks where additional diversity is
  interference. Error bars denote variation over three seeds.}
  \label{fig:real_robot_res}
  \vspace{-0.7em}
\end{figure}

We validate that the mastery--entropy behavior is not an artifact of
simulation by training ACT policies on the ARX arm across three tasks
(Fig.~\ref{fig:real_robot_res}); all results are averaged over three
seeds.

\textbf{StackCube: clear Inverted-U.}
On StackCube (Fig.~\ref{fig:real_robot_res}, left), success peaks at
$H_G \approx 3.3$. Below this peak, demonstrations are too geometrically
narrow to handle physical perturbations and pose variation; above it,
conflicting approach geometries increase ambiguity near contact, reducing
success—mirroring the low-to-mid mastery regime observed in simulation.

\textbf{PlacePanda and OpenDrawer: monotonic decline.}
For PlacePanda and OpenDrawer (Fig.~\ref{fig:real_robot_res}, center and
right), success decreases with $H_G$, consistent with higher-mastery
settings where low-entropy demonstrations already suffice and additional
geometric variation acts as noise rather than useful coverage.

\textbf{Sim-to-real consistency.}
The qualitative match—Inverted-U for the harder task and monotonic decay
for easier tasks—supports both the mastery--entropy principle and the use
of $H_G$ as a predictive pre-training diagnostic for real robot data.

% ─────────────────────────────────────────────────────────────────────
\subsection{Robustness of $H_G$: Overcoming Metric Aliasing}
\label{sec:metric_comparison}

\begin{figure}[tb]
      \centering
      \includegraphics[width=\columnwidth]{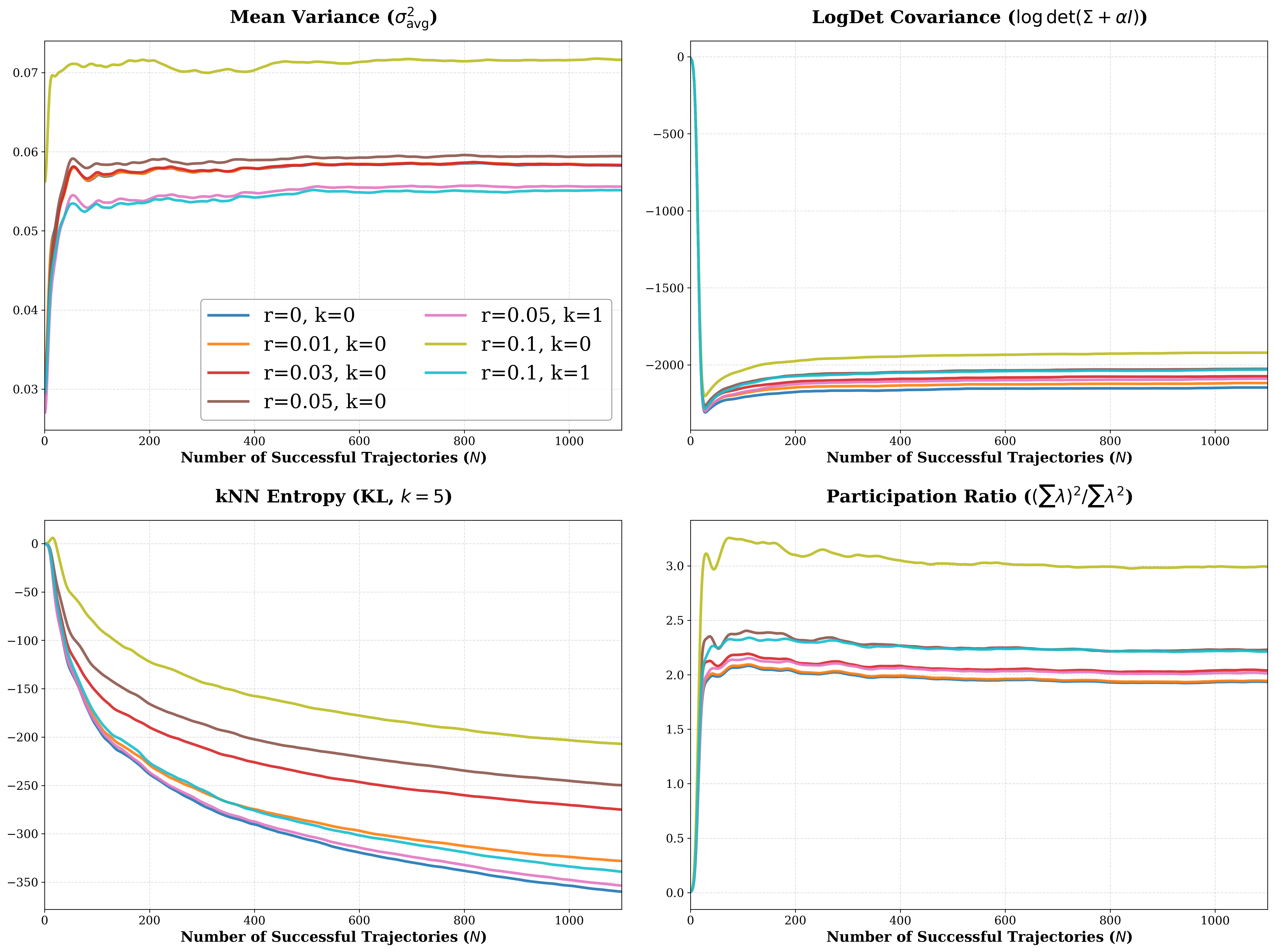}
      \caption{Baseline metric convergence on \texttt{StackCube-v1}. Mean
      variance, LogDet, and participation ratio converge quickly but
      collapse distinct $(r, k)$ settings (aliasing/mis-ordering), e.g.,
      $(0,0)$, $(0.01,0)$, and $(0.03,0)$ are nearly indistinguishable
      under variance despite increasing $H_G$. kNN entropy shows
      pronounced sample-size drift in the $D{=}150$ descriptor space.}
      \label{fig:metrics_comparison}
      \vspace{-0.7em}
\end{figure}

We test whether $H_G$ captures learning-relevant geometric structure
missed by common diversity proxies. We compare against four baselines,
all computed on the \emph{same aligned descriptors} as $H_G$
(Section~\ref{sec:alignment}): (1) \textbf{Mean Variance}
$\sigma^2_{\mathrm{avg}}$; (2) \textbf{LogDet Covariance}
$\log\det(\Sigma+\alpha I)$; (3) \textbf{Participation Ratio}
$(\sum_j \lambda_j)^2/\sum_j \lambda_j^2$; and (4) \textbf{kNN
Differential Entropy} (Kozachenko--Leonenko). We include $\alpha I$ only
to stabilize LogDet under $M \ll 3T$. Fig.~\ref{fig:metrics_comparison}
shows their behavior on \texttt{StackCube-v1} across the same $(r, k)$
configurations summarized by Table~\ref{tab:hg_values}.

\textbf{Baseline failure mode 1: marginal spread ignores mode organization (mean variance).}
$\sigma^2_{\mathrm{avg}}$ aggregates waypoint-wise spread and therefore
discards cross-time correlations and mode structure. As a result,
low-radius multi-mode settings $(r,k){=}(0,0)$, $(0.01,0)$, and $(0.03,0)$
collapse to nearly the same value despite increasing $H_G^{\mathrm{avg}}$
(Table~\ref{tab:hg_values}). Moreover, large-excursion single-mode
settings $(0.05,1)$ and $(0.10,1)$ can be mis-ordered relative to these
cases, since marginal spread alone cannot distinguish ``many tight modes''
from ``one consistent corridor.''

\textbf{Baseline failure mode 2: second-order summaries alias multimodality (LogDet and PR).}
LogDet and participation ratio incorporate spectral information but still
compress the dataset to a single covariance summary, so distinct
multimodal organizations can share similar values, leading to aliasing
across $(r,k)$. For example, participation ratio can place $(0.10,1)$
close to $(0.05,0)$ even though $H_G^{\mathrm{avg}}$ ranks $(0.10,1)$ much
lower, consistent with reduced strategy ambiguity under $k{=}1$. In
addition, LogDet requires explicit regularization under rank deficiency,
making it sensitive to $\alpha$.

\textbf{Baseline failure mode 3: instability at practical $M$ (kNN entropy).}
kNN differential entropy yields a broadly reasonable ordering but shows
pronounced sample-size drift over $N\le 1000$ in the $D{=}150$ descriptor
space, reducing its usefulness as a stable pre-training audit signal from
$O(10^2)$ demonstrations.

\textbf{Why $H_G$ is better behaved.}
In contrast, $H_G$ combines (i) the number of independent shape modes
(effective dimensionality; Eq.~\ref{eq:deff}) and (ii) their overall
spread (Eq.~\ref{eq:hg}), directly reflecting both multimodality and
geometric magnitude in aligned shape space. This makes it less prone to
collapsing distinct strategy organizations while converging rapidly
(Fig.~\ref{fig:convergence}).

%%%%%%%%%%%%%%%%%%%%%%%%%%%%%%%%%%%%%%%%%%%%%%%%%%%%%%%%%%%%%%%%%%%%%%%%%%%%%%%%
\section{DISCUSSION}
\label{sec:discuss}

\subsection{From Heuristic to Principle: Mastery--Entropy Matching}

Practitioners have long observed that heavily randomized demonstrations
can hurt imitation learning, but these effects are often treated as
idiosyncratic. Our results suggest a more general organizing principle:
a learner benefits from geometric diversity only up to a limit set by its
current \emph{task mastery}, beyond which additional modes introduce
ambiguity and degrade performance. This reframes ``data quality'' along
the trajectory-shape axis: instead of maximizing diversity by default,
one should match diversity to what the learner can reliably absorb.

% \subsection{Why Foundation Models May Need Cleaner Data}

% The VLA's monotonic decay with $H_G$ suggests that strong pretraining can
% shift the optimal diversity toward very low values. A plausible
% interpretation is that pretrained priors already provide broad geometric
% coverage, so fine-tuning diversity becomes redundant and can be
% interpreted as conflicting behavior. This implies that fine-tuning
% foundation models may require \emph{stricter} geometric consistency than
% training specialist policies from scratch.

\subsection{Limitations}
Our study focuses on a set of representative manipulation tasks in
simulation and on hardware. Extending the evaluation to a broader range
of task families (e.g., deformables, tool use, less contact-constrained
skills, and longer-horizon assembly) would help further assess how widely the mastery--entropy
behavior carries over and refine practical guidance across domains.

Our current pipeline leverages task semantics (e.g., gripper events) to
segment episodes into transit phases. Generalizing the phase
decomposition to settings without clear event markers---for example via
automatic segmentation or state-based heuristics---is a natural next
step.

Finally, the Inverted-U trend and peak-shift are empirical regularities
supported by our experiments. Developing a more formal treatment that
connects the optimal $H_G^*$ to factors such as model capacity,
conditioning information, and data coverage remains an interesting
direction for future work.

\section{CONCLUSION}
\label{sec:conclusion}

We introduced \textbf{Geometric Entropy} ($H_G$), a lightweight
pre-training metric for trajectory-shape diversity in robot demonstration
datasets. By aligning transit trajectories in a target-anchored frame to
remove extrinsic variation and summarizing the aligned shape manifold
with a stable spectral statistic, $H_G$ provides a practical audit signal
that can be computed before any policy training.

Across simulation and real-robot contact-rich manipulation tasks, we
observe a consistent \textbf{Inverted-U} relationship between success and
$H_G$: moderate diversity improves robustness via coverage, while
excessive diversity harms performance via mode ambiguity. The optimal
entropy $H_G^*$ shifts toward lower values as task mastery increases
through more data, easier tasks, or stronger priors, and can become
effectively monotonically decreasing for pretrained VLAs. Together, these
results challenge the ``maximize diversity'' heuristic and suggest a
calibration principle: match demonstration diversity to the learner's
mastery, using $H_G$ as a fast diagnostic for data curation.

% \addtolength{\textheight}{-12cm}

\section*{ACKNOWLEDGMENTS}
This work is supported by the Early Career Scheme of the Research Grants
Council (RGC) grant \# 27207224, the HKU-100 Award, a donation from the
Musketeers Foundation, and in part by the JC STEM Lab of Autonomous
Intelligent Systems funded by The Hong Kong Jockey Club Charities Trust.

\bibliographystyle{IEEEtran}
\bibliography{references}

\end{document}